\documentclass[sigplan,twocolumn]{acmart}
\renewcommand\footnotetextcopyrightpermission[1]{}
\AtBeginDocument{%
  }

\setcopyright{acmlicensed}
\copyrightyear{2018}
\acmYear{2018}
\acmDOI{XXXXXXX.XXXXXXX}
\acmConference[Conference acronym 'XX]{Make sure to enter the correct
  conference title from your rights confirmation email}{June 03--05,
  2018}{Woodstock, NY}
\acmISBN{978-1-4503-XXXX-X/2018/06}

\usepackage{tikz}
\usepackage{filecontents}

\usepackage{subfigure}
\usepackage{pifont}
\usepackage{amsmath, amsfonts}
\usepackage{multirow}
\usepackage{graphicx}
\usepackage{caption}
\usepackage{makecell}
\usepackage{pifont}
\usepackage{makecell}
\usepackage{tcolorbox}
\usepackage[table]{xcolor}

\usepackage{booktabs}

\usepackage{listings}
\usepackage{xcolor}

\usepackage{longtable}
\usepackage{ragged2e}

\usepackage[english]{babel}
\addto\extrasenglish{

}

\definecolor{attackgreen}{HTML}{56ad58}
\definecolor{attackred}{HTML}{aa260d}

\definecolor{codegreen}{rgb}{0,0.6,0}
\definecolor{codegray}{rgb}{0.5,0.5,0.5}
\definecolor{codepurple}{rgb}{0.58,0,0.82}
\definecolor{backcolour}{rgb}{0.95,0.95,0.92}
\definecolor{decoratorcolor}{rgb}{0.5,0,0.99} 

\lstdefinestyle{mypseudo}{
    backgroundcolor=\color{backcolour},
    basicstyle=\ttfamily\footnotesize,
    commentstyle=\color{codegreen},
    keywordstyle=\color{magenta}\bfseries,
    numberstyle=\tiny\color{codegray},
    breaklines=true,
    captionpos=b,
    keepspaces=true,
    columns=flexible,
    numbersep=5pt,
    showstringspaces=false,
    showtabs=false,
    tabsize=2,
    frame=single,
    language={},
    morekeywords={
        Kernel, for, each, do, end, while,
        STORE, LOAD, L2_INVALIDATE,
        YIELD_TO_VICTIM, COUNT
    },
    morecomment=[l]{\#}
}

\begin{document}

\title{\textsc{SparLeak}: Privacy Leakage from Sparse Attention in LLM Inference on Shared GPUs}
\author{Fahao Chen}
\email{chenfh@ieee.org}
\affiliation{%
  \institution{Shandong University}
  \country{China}
}

\author{Linkang Du}
\affiliation{%
  \institution{Xi’an Jiaotong University}
  \country{China}}
\email{linkangd@xjtu.edu.cn}

\author{Jinhao Zhou}
\affiliation{%
  \institution{Waseda University}
  \country{Japan}}
\email{jinhao@suou.waseda.jp}

\author{Peng Li}
\affiliation{%
  \institution{Xi’an Jiaotong University}
  \country{China}}
\email{pengli@xjtu.edu.cn}

\author{Zhou Su}
\affiliation{%
 \institution{Xi’an Jiaotong University}
 \country{China}}
\email{zhousu@ieee.org}

\renewcommand{\shortauthors}{Trovato et al.}

\begin{abstract}
  Sparse attention is widely used to accelerate long-context inference in modern large language models (LLMs), but its input-dependent execution behavior introduces previously unexplored privacy risks. We identify a new GPU micro-architectural side channel, termed Sparsity-Induced Memory Access (SIMA), which arises from secret-dependent key-value cache access patterns induced by sparse attention. 

  Based on this observation, we present \textsc{SparLeak}, a phase-aware side-channel attack that extracts SIMA traces during LLM inference and enables two practical privacy extractions: query attribute inference from prefill-phase traces and autoregressive response reconstruction from decoding-phase traces. By reconstructing approximate token-level sparsity profiles from page-level observations and applying profiling-based learning, \textsc{SparLeak} accurately recovers sensitive information, including user-query attributes and private LLM response content. Extensive evaluation across three LLM architectures, three sparse attention mechanisms, and three privacy-sensitive datasets shows that \textsc{SparLeak} achieves average attack success rates of 90.9\% for attribute inference and 87.3\% for response reconstruction under real-world LLM serving settings, highlighting the significance to account for SIMA leakage when deploying sparse-attention-based LLM systems.
  We provide anonymized SIMA traces, trained attack models, evaluation scripts, and documentation as artifacts at \href{https://anonymous.4open.science/r/Janus_artifacts/}{https://anonymous.4open.science/r/SparLeak\_artifacts/}.
\end{abstract}





\pagestyle{plain}

\maketitle

\section{Introduction}

The widespread deployment of large language models (LLMs) \cite{touvron2023llama,team2024gemma} has driven the use of increasingly long context to improve generation quality and reduce hallucinations. However, the quadratic cost of attention makes long-context inference computationally expensive. To mitigate this overhead, modern LLM systems increasingly adopt \textit{sparse attention} mechanisms~\cite{zhang2023h2o,tang2024quest,cho2024sparc}, which approximate full attention by selectively attending to a subset of tokens. This trend is reflected in long-context serving systems such as SGLang~\cite{zheng2024sglang} and vLLM~\cite{kwon2023efficient}, and is likely to intensify as model scales and context lengths grow. The tokens involved in sparse attention computation are typically identified using activation-based saliency~\cite{zhang2023h2o, cho2024sparc, desaihashattention, xu2025xattention}, block-level scoring~\cite{tang2024quest}, or learned predictors~\cite{treviso2022predicting, gao2024seerattention, xu2025specontext}. By pruning irrelevant tokens, sparse attention significantly reduces both prefill-phase and decoding-phase costs.

While designed for efficiency, sparse attention poses a new privacy threat in LLM inference. We find that sparse attention induces secret-dependent memory accesses in the GPU memory hierarchy by dynamically selecting and retrieving Key-Value (KV) tensors based on semantic relevance. These accesses occur internally during GPU execution and are not directly exposed through the LLM service interface. However, they induce observable contention in shared GPU micro-architectural components, such as caches and TLBs. From a security perspective, this effect gives rise to a new side-channel surface rooted in sparsity-induced memory access (SIMA) behavior, which has not been explored by existing works. The SIMA side channel captures how sparse attention externalizes semantic information through data-dependent interactions with the GPU memory hierarchy. 

Compared to prior side channels (in \autoref{tab:practicality_comparison}) for exploring privacy leakage during LLM inference based on timing~\cite{zhang2024time, song2025early, wu2025know, zheng2024inputsnatch}, network traffic~\cite{weiss2024your, mcdonald2025whisper}, or embedding cache behavior~\cite{adiletta2025spill, gao2025know}, we find that SIMA side-channel surface is model-intrinsic and spans both prefill and decoding phases, thereby enabling broader and more persistent privacy leakage. These properties reveal an emerging privacy risk inherent to sparse attention mechanisms and motivate timely investigation in modern LLM deployments~\cite{sun2025efficient}.


\begin{table*}[t]
\centering
\small
\setlength{\tabcolsep}{6pt}
\resizebox{\linewidth}{!}{
\begin{tabular}{l|cccccc}
\hline
\multirow{2}{*}{\textbf{Side-Channel Attacks}}
& \multirow{2}{*}{\textbf{Threat Surface}}
& \multirow{2}{*}{\textbf{Extracted Info.}}
& \multirow{2}{*}{\textbf{\makecell[c]{Prefill-Phase\\Attack}}}
& \multirow{2}{*}{\textbf{\makecell[c]{Decoding-Phase\\Attack}}}
& \multicolumn{2}{c}{\textbf{Mitigation}}\\
& & & & & \textbf{Prevention} & \textbf{Detection}\\
\hline
Time Will Tell~\cite{zhang2024time} & Timing (response time) & Response length & \ding{52} & \ding{56} & \ding{56} & \ding{56} \\

Early Bird~\cite{song2025early} & Serving order & KV Cache hit/miss & \ding{52} & \ding{56} & \cite{chu2025selective, luo2025shadow} & \cite{chu2025selective}\\

PromptPeek~\cite{wu2025know} & Timing (time-to-first-token) & KV Cache hit/miss & \ding{52} & \ding{56} & \cite{chu2025selective, luo2025shadow} & \cite{chu2025selective} \\

InputSnatch~\cite{zheng2024inputsnatch} & Timing (time-to-first-token) & KV Cache hit/miss\&hit ratio & \ding{52} & \ding{56} & \cite{chu2025selective, luo2025shadow} & \cite{chu2025selective} \\

What Was Your Prompt~\cite{weiss2024your} & Network traffic & Token-length sequence & \ding{52} & \ding{52} & \cite{mehta2022pacer, zhang2025netecho} & \ding{56}\\

Whisper Leak~\cite{mcdonald2025whisper} & Network traffic\&timing & Token-length sequence\&timing sequence & \ding{52} & \ding{56} & \cite{mehta2022pacer, zhang2025netecho} & \ding{56} \\

Spill The Beans~\cite{adiletta2025spill} & CPU shared cache (embedding table) & Embedding access hits & \ding{56} & \ding{52} & \cite{wang2019cacheguard} & \cite{lamster2025waitwatcher}\\

I Know What You Said~\cite{gao2025know} & CPU shared cache (embedding table) & Embedding access hits\&access order  & \ding{52} & \ding{52} & \cite{wang2019cacheguard} & \cite{lamster2025waitwatcher}\\

\rowcolor{gray!30}
\textbf {This work} & Sparsity-induced memory access & Sparsity patterns & \ding{52} & \ding{52} & \ding{56} & \ding{56}\\
\hline
\end{tabular}
}
\caption{Comparison of representative side-channel attacks that expose privacy leakage in LLM serving. \textbf{Prefill-Phase Attack} and \textbf{Decoding-Phase Attack} indicate whether an attack can infer sensitive query information or recover generated outputs, respectively. \textbf{Mitigation} summarizes existing defense efforts: entries with citations indicate that prior prevention or detection mechanisms have been proposed and can be applied to mitigate the corresponding attack, while \ding{56} denotes that no effective mitigation has been reported to date.}
\label{tab:practicality_comparison}
\end{table*}

\noindent\textbf{Challenges.} While the SIMA side channel exposes a promising and principled attack surface, translating this leakage into practical, end-to-end attacks is non-trivial. Successfully exploiting SIMA in real LLM deployments requires overcoming several critical technical challenges. 
\underline{First}, SIMA behavior is not directly observable. An attacker cannot see KV accesses or sparsity decisions, and must instead infer them indirectly from noisy GPU micro-architectural signals. Compounding this difficulty, LLM inference consists of two execution phases, prefill and decoding, with fundamentally different parallelism and access patterns: the prefill phase processes tokens in parallel and thus exposes a cumulative signal, whereas the decoding phase proceeds token by token and exposes a step-wise signal~\cite{zhong2024distserve, patel2024splitwise}. These phases produce SIMA traces with distinct temporal structures and leakage characteristics, making it necessary to design phase-specific extraction methods.
\underline{Second}, even when micro-architectural leakage can be captured, it is spatially coarse. KV tensors are managed at virtual-memory page granularity~\cite{kwon2023efficient}, and each observed page access aggregates the KV states of many tokens. Consequently, the attacker only observes page-level SIMA traces that collapse multiple token-level accesses into a single event, obscuring fine-grained sparsity structure and discarding much of the information needed to distinguish individual tokens or prompt regions.
\underline{Third}, sparse attention traces encode computational importance rather than explicit lexical content. Recovering meaningful private information therefore requires reconstructing latent access structure and learning the statistical relationship between SIMA traces and high-level semantics or token identities.

\noindent\textbf{Solutions.} We present \textsc{SparLeak}, the first phase-aware side-channel framework that exploits SIMA behavior during LLM inference. \textsc{SparLeak} observes GPU micro-architectural signals and abstracts them into SIMA traces that reflect the input-dependent KV access behavior in prefill and decoding phases, respectively. \textsc{SparLeak} then converts these traces into approximate sparsity patterns and uses them as a structural signal to infer sensitive information without accessing the victim's KV contents. To enable practical exploitation of SIMA leakage, \textsc{SparLeak} incorporates the following designs.


First, \textsc{SparLeak} introduces two GPU side-channel primitives to extract SIMA traces. It distinguishes prefill and decoding phases solely from GPU execution dynamics: the prefill phase performs highly parallel computation over the input context, whereas the decoding phase proceeds sequentially, generating one token at a time. Exploiting this transition, \textsc{SparLeak} recovers cumulative SIMA traces during the prefill phase using an L2-cache–based \textsc{Invalidate+Compare} primitive, and extracts step-wise SIMA traces during the decoding phase using a TLB-based \textsc{Evict+Reload} primitive.

Second, to overcome spatial aggregation at page granularity, \textsc{SparLeak} performs trace reconstruction to infer approximate token-level sparsity profiles. Rather than attempting exact recovery, \textsc{SparLeak} exploits spatial locality and distributional regularities in sparse attention. By enforcing statistical consistency with observed SIMA traces, it reconstructs latent access structure sufficient to restore discriminative power for downstream inference.

Finally, \textsc{SparLeak} treats reconstructed SIMA traces as a latent structural signal and leverages their statistical correlation with semantic attributes and token identities. Specifically, \textsc{SparLeak} learns a model that maps from reconstructed traces to private information through offline profiling. Based on this principle, \textsc{SparLeak} realizes two end-to-end inference attacks: Query Attribute Inference (QAI), which infers high-level semantic properties of user prompts from prefill-phase traces, and Autoregressive Token Recovery (ATR), which reconstructs generated tokens from decoding-phase traces. In both cases, the adversary applies a learned mapping at runtime without access to the victim’s model, prompts, or outputs.


\noindent\textbf{Evaluation.} We conduct an extensive evaluation of \textsc{SparLeak} across multiple widely-adopted LLMs, sparse attention mechanisms, and privacy-sensitive datasets. The results demonstrate that SIMA constitutes a powerful side channel, enabling accurate attack of both query attributes and generated responses. Across a broad range of configurations, \textsc{SparLeak} achieves average attack success rates of 90.9\% for query attribute inference and 87.3\% for autoregressive token recovery. Notably, even under realistic system noise and model evolution, \textsc{SparLeak} retains over 90\% of its original attack effectiveness. We further evaluate a potential mitigation by injecting input-independent randomness into sparsity patterns. While this approach reduces attack effectiveness, it does so at the cost of degraded inference quality.
\section{Background and Motivation}
\label{sec:bc_moti}

In this section, we first give background about LLM inference with sparse attention, followed by the motivation of privacy leakage in sparse attention. 
\subsection{Large Language Models Inference}
\noindent\textbf{Attention Mechanism.} Modern LLMs are primarily built on the Transformer architecture~\cite{vaswani2017attention}, whose core component is multi-head self-attention. During inference, LLM execution consists of two phases: \emph{prefill} and \emph{decoding}. In the prefill phase, the model processes the entire input sequence in parallel, with computation and memory cost scaling with input length. In the decoding phase, each newly generated token attends to all previously processed tokens. To avoid redundant computation, modern inference engines maintain a key-value (KV) cache that stores the $K$ and $V$ representations of past tokens and reuses them across decoding steps~\cite{kwon2023efficient, lee2024infinigen}.


\noindent\textbf{Sparse attention.} To improve inference efficiency, recent work proposes sparse attention mechanisms~\cite{zhang2023h2o, tang2024quest, xu2025xattention, desaihashattention, treviso2022predicting}, motivated by the observation that only a small subset of tokens contributes substantially to attention outputs. Rather than attending to all tokens, sparse attention selectively restricts attention computation to a subset of tokens chosen based on the current query or decoding token, reducing computation and memory overhead in both prefill and decoding phases.

Existing sparse attention approaches mainly focus on identifying important tokens without explicitly computing full attention scores. They can be broadly categorized into three classes: 
(1) \emph{activation-based saliency} methods that infer importance from model activations or residual norms~\cite{zhang2023h2o, tang2024quest, xu2025xattention, cho2024sparc}; 
(2) \emph{block-level scoring} methods that approximate attention at the block or page level and retain tokens from high-scoring blocks~\cite{desaihashattention}; and 
(3) \emph{learned predictors} that use lightweight models to directly predict important tokens~\cite{treviso2022predicting, gao2024seerattention, xu2025specontext}.
\begin{figure}[t]
    \centering
    \includegraphics[width=\linewidth]{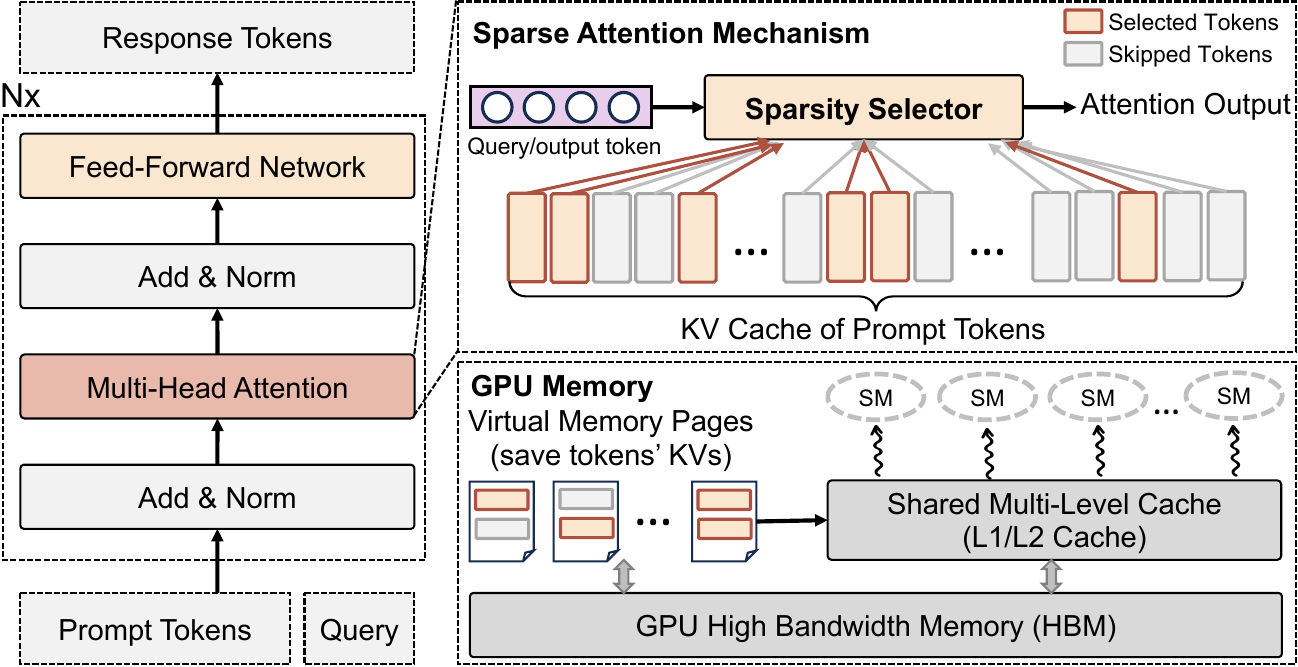}
    \caption{The illustration of sparse attention mechanism in LLM inference and its associated GPU memory access.}
    \label{fig:bc}
\end{figure}

\noindent\textbf{GPU memory access behavior.} As shown in \autoref{fig:bc}, the KV cache of tokens is stored in GPU high-bandwidth memory (HBM) and managed by the GPU virtual memory system, which operates at page granularity~\cite{allen2021depth, nazaraliyev2024gpuvm}. On contemporary NVIDIA GPUs, virtual memory is managed using fixed-size pages that span 64~KB to 2~MB in contiguous virtual address space~\cite{allen2021depth}. In contrast, prior work shows that the KV tensors associated with a single token occupies only a few kilobytes per layer, typically in the range of 10-30~KB~\cite{kwon2023efficient}. As a result, a single GPU memory page typically co-locates the KV tensors of multiple tokens.

Under sparse attention, the model first identifies a subset of important tokens for a given token. This selection is ultimately realized at page granularity by the GPU memory subsystem: when a token is selected, the virtual memory pages containing its KV tensors are fetched from HBM into the on-chip cache hierarchy~\cite{dao2022flashattention}. The fetched KV tensors are then consumed by streaming multiprocessors (SMs) for attention computation, while tokens that are not selected do not trigger page accesses. Consequently, sparse attention induces data-dependent page-level memory access patterns that implicitly aggregate the activities of multiple tokens co-located within each page.

\subsection{Motivation}
In this section, we characterize sparsity patterns induced by sparse attention during LLM inference. These patterns differ between the prefill and decoding phases, revealing sensitive information about user queries and LLM generated responses, which motivates the SIMA side channel.

\begin{tcolorbox}[colback=gray!10,colframe=black]
\noindent\textbf{Motivation 1:} \textit{Sparse attention induces distinctive access patterns in the prefill and decoding phases.}
\end{tcolorbox}
\begin{figure}[t]
    \centering
    \includegraphics[width=\linewidth]{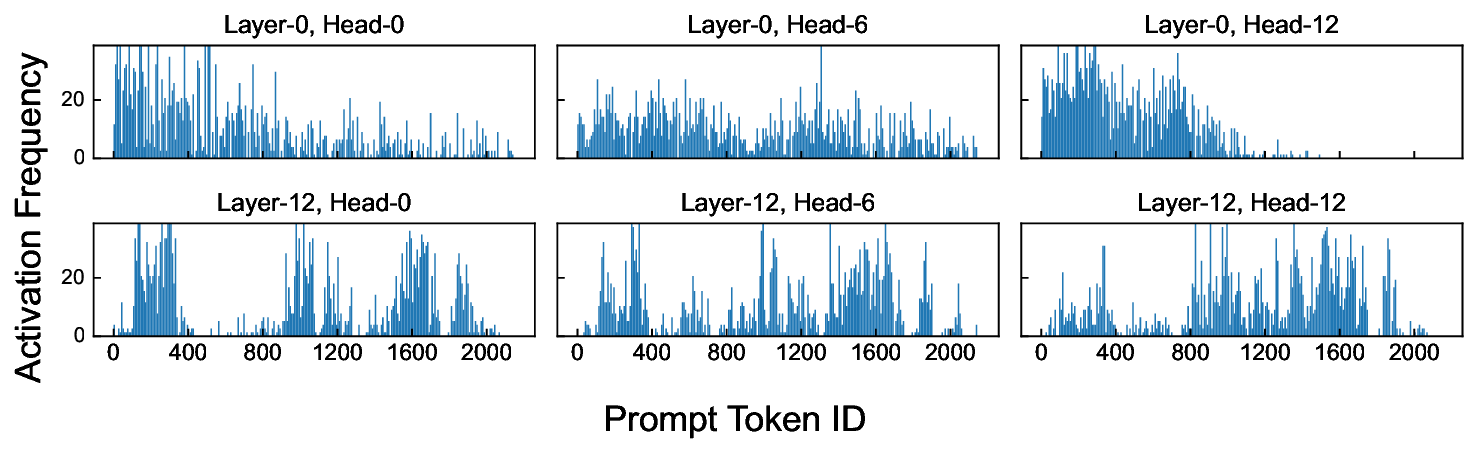}
    \caption{Sparsity pattern in prefill phase. Activation frequency denotes how often a given token is selected by sparse attention during the prefill phase.}
    \label{fig:prefil_sparsity}
\end{figure}
Sparse attention mechanisms~\cite{zhang2023h2o,tang2024quest,cho2024sparc} fundamentally alter the execution behavior of LLM inference. By selectively computing attention over only a subset of tokens, sparse attention induces structured, input-dependent selection of key-value (KV) states, in contrast to dense attention where all tokens are uniformly accessed. While these selection decisions are made at the model level, they directly determine which KV tensors are accessed during inference and how frequently they are reused. From an execution perspective, such non-uniform KV access behavior inevitably translates into data-dependent interactions with the GPU memory hierarchy. Therefore, we characterize the sparsity patterns produced by sparse attention in the prefill and decoding phases, and show that they exhibit qualitatively different structures across the two phases. These differences motivate a phase-aware view of sparsity-induced leakage and SIMA-based side-channel attacks.

We use the LLaMA3-8B~\cite{grattafiori2024llama} model with sparse attention enabled. In our setting, each inference request is formed by concatenating a fixed system prompt and a user-provided query. Unless otherwise stated, the prompt is identical across all motivation experiments and is constructed from an open-source healthcare dataset~\cite{health}. All inputs are issued under a fixed system prompt built from an open-source healthcare dataset, with a total prompt length of approximately 2K tokens. During the prefill phase, sparse attention selects the top 10\% of prompt tokens for each query token based on approximate attention scores~\cite{desaihashattention}.



\noindent\textbf{Prefill-phase sparsity pattern.}
During the prefill phase, LLM processes entire input context in parallel under sparse attention. Rather than attending uniformly to all context tokens, sparse attention selectively activates a subset of tokens whose KV states contribute to the attention computation. As prefill proceeds, these selective activations form a characteristic sparsity pattern over the input context, reflecting how attention is distributed across different parts of the input.

We visualize this sparsity pattern for an example input in \autoref{fig:prefil_sparsity}. Two observations stand out. First, attention activation is highly skewed: a small subset of tokens is consistently emphasized, while most tokens receive little or no attention. Second, sparsity patterns vary substantially across attention layers and heads. As shown in \autoref{fig:prefil_sparsity}, different layers and heads focus on different parts of the input, suggesting that sparse attention captures complementary semantic information across model components.


\begin{figure}[t]
    \centering
    \includegraphics[width=\linewidth]{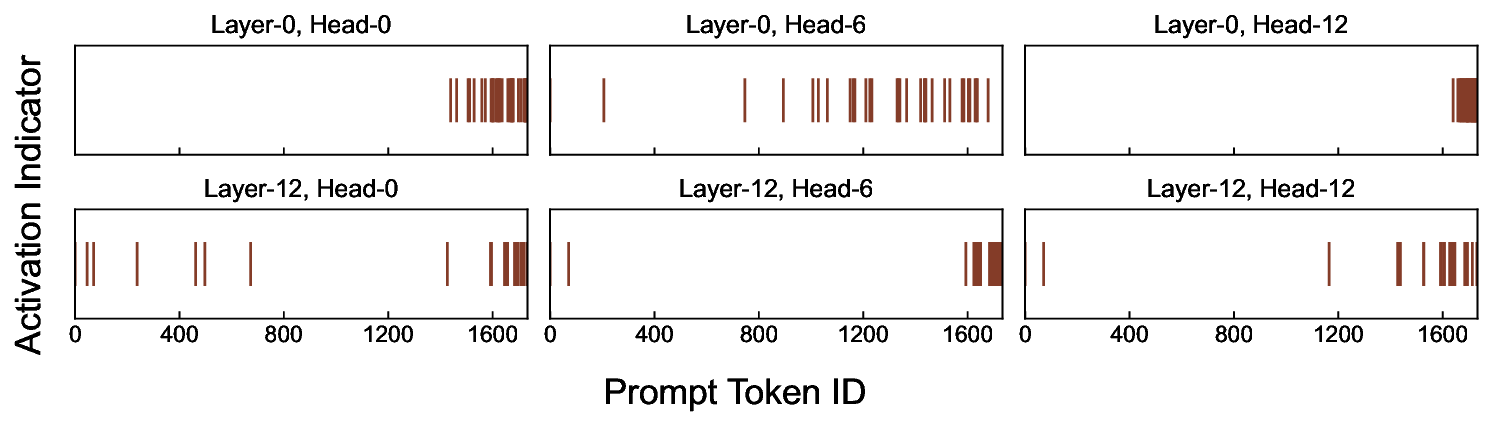}
    \caption{Sparsity pattern at a selected decoding step. Activation indicator is a binary value indicating whether a token is selected by sparse attention at the given decoding step.}
    \label{fig:decoding_sparsity}
\end{figure}
\noindent\textbf{Decoding-phase sparsity pattern.}  
In contrast to the prefill phase, decoding phase proceeds autoregressively. At each decoding step, sparse attention selectively activates only a limited subset of context tokens when computing the next-token representation. Consequently, the sparsity pattern observed at a given step reflects which parts of the input context are selectively attended at that moment.

We illustrate this behavior by plotting the sparsity pattern at a representative decoding step in \autoref{fig:decoding_sparsity}. As shown, each decoding step activates only a small fraction of the context, confirming that decoding-phase sparsity is highly selective and fine-grained. Similar to the prefill phase, the activated tokens differ markedly across layers and heads, resulting in heterogeneous yet structured sparsity patterns across the LLM. 


\begin{tcolorbox}[colback=gray!10,colframe=black]
\noindent\textbf{Motivation 2:} \textit{Prefill- and decoding-phase sparsity patterns correlate with sensitive query and responses.}
\end{tcolorbox}
\begin{figure}[t]
    \centering
    \includegraphics[width=\linewidth]{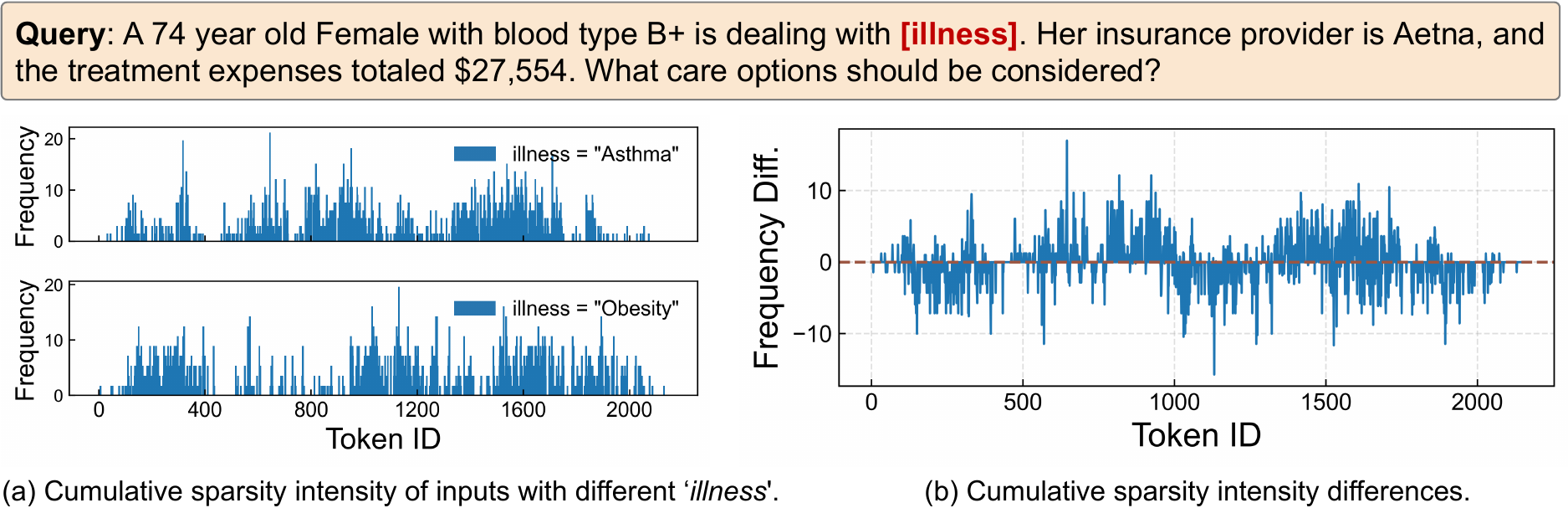}
    \caption{Sparsity patterns of user queries with different attributes. In this example, we change the illness categories in different queries.}
    \label{fig:moti_2}
\end{figure}

Building on our analysis of attention sparsity patterns, we first examine whether the sparsity induced during the prefill phase correlates with high-level semantic attributes of user inputs. This question is motivated by prior work showing that side-channel signals derived from model execution can support attribute-level inference even without reconstructing the exact input content~\cite{lukas2023analyzing,kim2023propile,liu2025evaluating}.

\autoref{fig:moti_2} compares prefill-phase sparsity patterns for two inputs that differ only in a single semantic attribute, namely the illness category (e.g., \textit{Obesity} versus \textit{Asthma}), while all other attributes and the system prompt remain identical. As shown in \autoref{fig:moti_2}(a), the two inputs induce noticeably different sparsity distributions over the same prompt. Certain parts of the prompt are consistently emphasized under one illness category but not the other. To further characterize this effect, \autoref{fig:moti_2}(b) visualizes the difference between the two sparsity patterns, revealing systematic and non-negligible deviations across the prompt.

These results indicate that prefill-phase sparsity patterns are sensitive to high-level semantic attributes of the input. Even when the overall prompt structure and length are fixed, modifying a single semantic attribute reshapes which parts of the prompt are repeatedly attended and how strongly. This observation motivates our \emph{Query Attribute Inference} attack: if an adversary can recover sparsity information corresponding to the prefill phase, it may infer sensitive semantic properties of the user input without reconstructing the exact prompt text.




\begin{figure}[t]
    \centering
    \includegraphics[width=\linewidth]{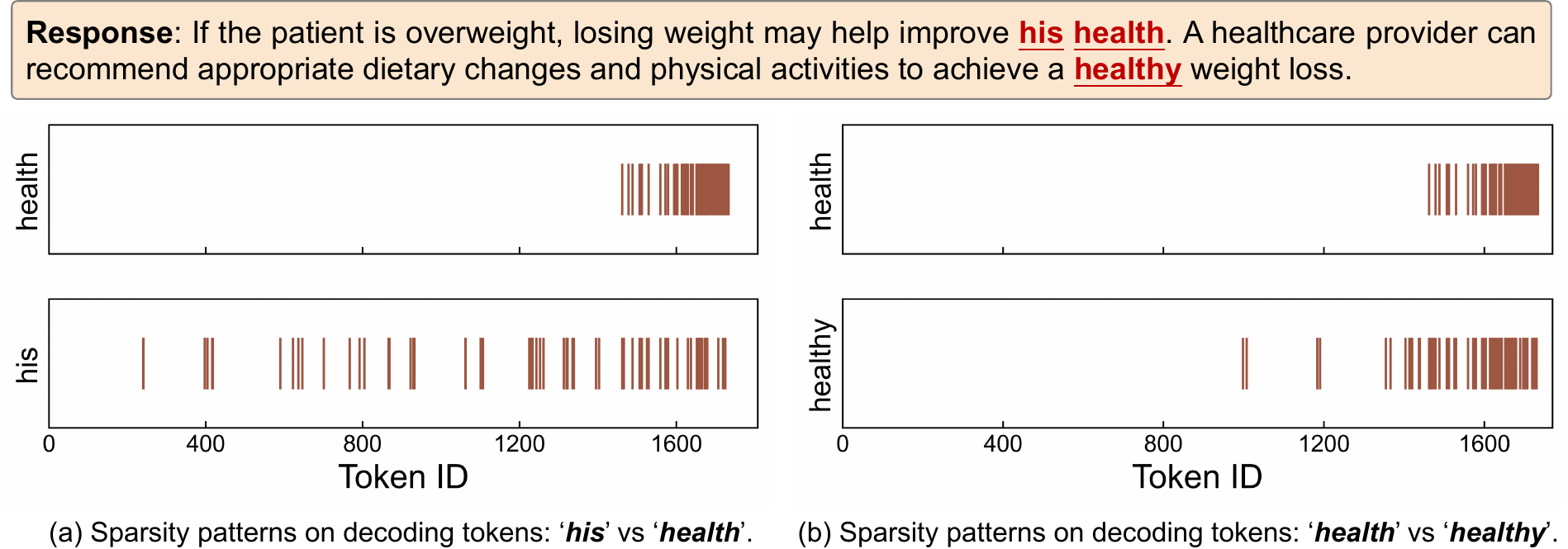}
    \caption{Sparsity patterns of different decoding tokens, including ``his'', ``health'', and ``healthy''.}
    \label{fig:moti_3}
\end{figure}

We then examine whether sparsity patterns during the decoding phase encode information about generated responses. Using the same LLM, prompt, and sparsity configuration, we analyze the attention sparsity pattern at each decoding step. At a given step, the pattern reflects which parts of the prompt are attended when computing the next output token. \autoref{fig:moti_3} visualizes representative decoding-phase sparsity patterns induced by different output tokens. As shown in \autoref{fig:moti_3}(a), distinct tokens (e.g., \textit{``his''} versus \textit{``health''}) lead to clearly different sparsity patterns over the prompt. Notably, even semantically related tokens (e.g., \textit{``health''} and \textit{``healthy''}) can induce distinguishable sparsity patterns, as illustrated in \autoref{fig:moti_3}(b). This arises because sparse attention is conditioned not only on the current token but also on the previously generated context, which evolves over the decoding phase and reshapes the attention distribution.

Taken together, these observations show that decoding-phase sparsity patterns are strongly token-dependent. Each generated token induces a distinctive attention footprint shaped jointly by its semantic content and its position in the response sequence. This property underpins our \emph{Autoregressive Token Recovery} attack: if an adversary can infer sparsity patterns across decoding steps, it may progressively distinguish candidate tokens and reconstruct the generated response.

\section{Attack Overview}

This section presents the threat model, SIMA leakage mechanism, and end-to-end attack workflow.

\subsection{Threat Model}
\begin{figure}[t]
    \centering
    \includegraphics[width=\linewidth]{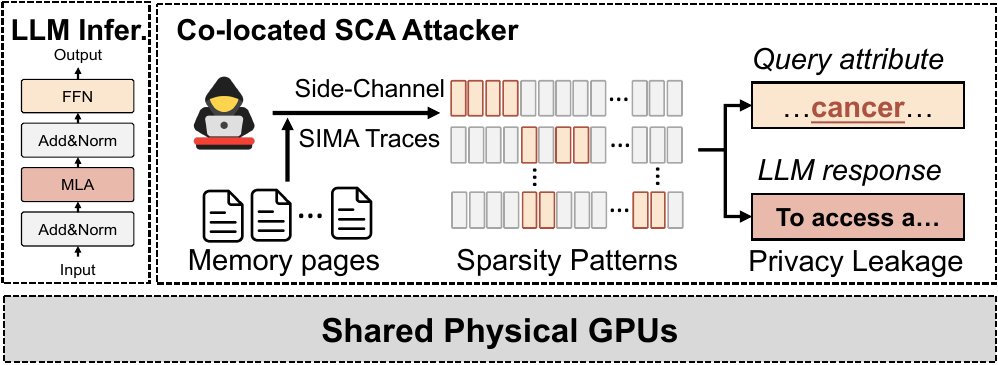}
    \caption{The illustration of the threat model.}
    \label{fig:threa_model}
\end{figure}

As shown in \autoref{fig:threa_model}, we consider a realistic LLM serving deployment in which a victim runs sparse-attention-enabled inference on a modern GPU. Specifically, we target a service-style setting in which the victim application uses a public and fixed system-prompt template across requests, while the user query and generated response remain private, as in open-source RAG and agent applications that expose and reuse standardized templates, often with prefix caching. This setting reflects current practice in long-context LLM serving, where sparse attention is increasingly adopted to reduce overhead~\cite{zhang2023h2o, tang2024quest, xu2025xattention, desaihashattention}.  


\noindent\textbf{Adversary and attack goal.}  
We consider an unprivileged adversary that runs a co-resident spy process on the same physical GPU as the victim, following the standard co-residency threat model in GPU side-channel attacks~\cite{zheng2024inputsnatch, umar2025efficient}. The spy process issues GPU probing operations to observe micro-architectural effects induced by the victim's sparse-attention computation.

The adversary’s goal is to infer sensitive information about both the user query and the LLM's response via the SIMA side channel. Concretely, the adversary (i) performs \emph{Query Attribute Inference} (QAI) by exploiting cumulative SIMA behavior during the prefill phase, and (ii) conducts \emph{Autoregressive Token Recovery} (ATR) by leveraging step-wise SIMA observations during the decoding phase.


\noindent\textbf{Adversary capabilities.}  
Consistent with prior work~\cite{gao2025know, umar2025efficient}, we consider an unprivileged adversary co-resident with the victim on the same physical GPU. This setting can arise when a local user runs a CUDA workload on a shared GPU or when a remote tenant's job is co-scheduled on the same non-partitioned GPU. The adversary requires only user-level CUDA access and monitors memory-related micro-architectural effects, including L2 cache activity and TLB residency across context switches, by measuring the latency of its own probes. This requires no privileged counters, driver modifications, or access to victim memory. The resulting observations expose secret-dependent access patterns induced by sparse attention without directly revealing KV contents or exact victim addresses.

In addition, the adversary has no control over GPU scheduling or exact execution timing. Under the fixed-template serving setting and a fixed model and runtime configuration, we further assume that the adversary can calibrate contention-based probe mappings for the corresponding prompt-side KV pages offline and reuse them across victim invocations, without accessing KV contents or recovering exact runtime addresses.


\noindent\textbf{Threat surface and practicality.} Data-dependent KV accesses induced by sparse attention create the observable GPU memory effects exploited by \textsc{SparLeak}. We consider co-resident execution alongside deployed LLM services during normal, uninterrupted operation~\cite{zheng2024sglang}. Such services maintain long-lived model processes to avoid repeated initialization and continuously serve incoming requests. In this setting, the monitored prompt-side KV allocations are reused across requests, allowing offline page-probe mappings to remain applicable.



\begin{figure*}[t]
    \centering
    \includegraphics[width=\linewidth]{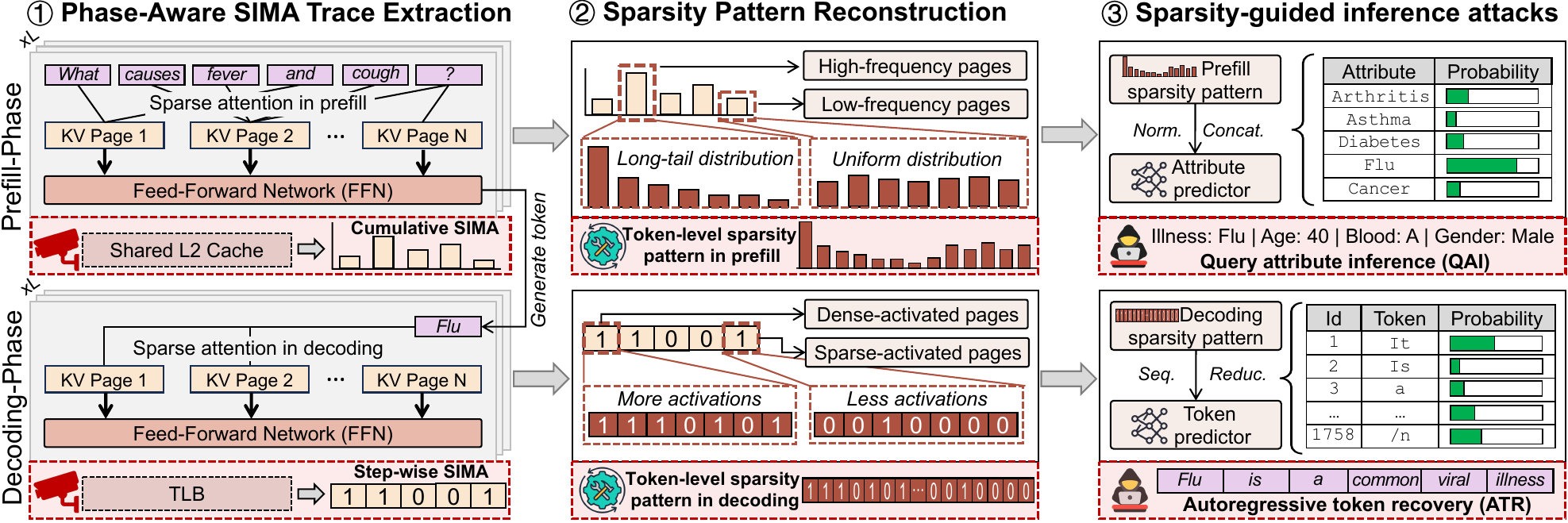}
    \caption{The overview of the attack workflow. \textsc{SparLeak} proceeds in three stages: (1) phase-aware extraction of SIMA traces for prefill and decoding phases using different side-channel primitives, (2) reconstruction of token-level sparsity patterns from page-level SIMA traces, and (3) sparsity-guided inference for query attribute inference and autoregressive token recovery.}
    \label{fig:overview}
\end{figure*}


\subsection{Attack Workflow}
As shown in \autoref{fig:overview}, \textsc{SparLeak} follows a three-stage attack pipeline that progressively transforms low-level GPU micro-architectural observations into high-level privacy leakage. 
\ding{192} \ding{192} First, it extracts phase-aware SIMA traces from L2 cache and TLB activity, using the KV-cache layout and calibrated probe mappings to distinguish access features across layers and KV heads. It exploits the distinct execution characteristics of prefill and decoding to obtain cumulative traces for the prefill phase and step-wise traces for the decoding phase.
\ding{193} Second, it reconstructs approximate token-level sparsity patterns from coarse page-level SIMA traces by leveraging the locality and statistical regularity inherent in sparse attention, enabling probabilistic refinement despite page-level aggregation. 
\ding{194} Finally, it performs sparsity-guided inference attacks that incorporate phase-specific leakage semantics, by training lightweight MLP-based classifiers on the reconstructed sparsity patterns, enabling query attribute inference during the prefill phase and autoregressive token recovery during the decoding phase. 
\section{Phase-Aware SIMA Trace Extraction}\label{sec:sima_extraction}
In this section, we first describe how to distinguish the prefill and decoding phases, and then introduce two primitives for SIMA trace extraction in the prefill and decoding phases, respectively.

\subsection{Prefill and Decoding Distinction}
We distinguish the \textit{prefill} and \textit{decoding} phases based on their distinct GPU execution patterns. During the prefill phase, the model processes the entire input context in a single forward pass, triggering intensive matrix computations over the full context length. Consequently, the prefill phase exhibits a short but concentrated burst of compute-heavy GPU kernels, resulting in high kernel-time density. Here, a kernel refers to a computation routine executed on the GPU, such as matrix multiplication or attention computation. In contrast, the decoding phase generates tokens iteratively, with each token requiring an incremental forward pass. This produces a long execution trace with a near-periodic structure corresponding to successive token generations.

To locate the transition between these temporal patterns, one feasible approach uses the kernel timeline from Nsight Systems~\cite{nsight} as an independent boundary reference. Nsight was accessible without root privileges in our setup. Using this timeline, GPU activity can be quantified by a kernel-time-based utilization proxy:
$\textit{util}(w) = \frac{\tau}{|w|}$,
where $\tau$ denotes the cumulative kernel execution time within a fixed window $w$, and $|w|$ is the window length. Focusing on \texttt{GEMM} kernels provides a cleaner and more stable signal, as they dominate LLM computation. The same transition from concentrated activity to near-periodic execution can also be inferred directly from the intensity and periodicity of the collected L2/TLB traces.


\subsection{SIMA Trace Extraction} \label{sec:trace_extraction}

\subsubsection{Prefill-Phase SIMA Trace Extraction} \label{sec:sparsity_prefill}
During this phase, the LLM processes the prompt and user query in parallel. Under sparse attention, query tokens selectively access KV states associated with prompt tokens. We define the prefill-phase leakage as a \emph{cumulative SIMA trace}, denoted by $\{c_{1}, c_{2}, \ldots, c_{N}\}$, where $N$ is the number of monitored prompt-side KV pages. Each $c_i$ estimates how frequently page $p_i$ is accessed across query-token attention computation during the prefill phase.

\noindent\textbf{Attack Implementation.} \textsc{SparLeak} leverages the observation that repeated sparse accesses to prompt-side KV pages create measurable replacement pressure in the shared GPU L2 cache. Although the exact GPU L2 replacement policy is undocumented, prior work shows that cache contention produces stable eviction behavior~\cite{zhang2024invalidate+}. During offline calibration, for each monitored page $p_i$, the attacker constructs a small set of cache-line-aligned probe addresses $s_i$ whose L2 eviction behavior is correlated with accesses to $p_i$, following the methodology of~\cite{zhang2024invalidate+}.

\begin{figure}[t]
    \centering
    \includegraphics[width=\linewidth]{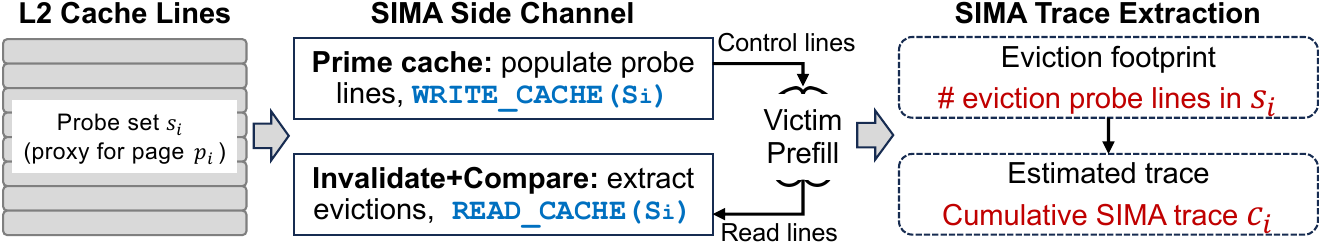}
    \caption{Spy process for estimating page-level access frequency during the prefill phase by repeatedly measuring cache replacement effects induced by sparse-attention KV accesses.}
    \label{fig:prefill_attack}
\end{figure}

As illustrated in \autoref{fig:prefill_attack}, at runtime the spy invokes \texttt{WRITE\_CACHE($s_i$)} to populate the calibrated probe lines in L2 and applies \texttt{READ\_CACHE($s_i$)} with the \textsc{Invalidate}+\textsc{Compare} primitive while the victim executes the prefill phase under default GPU scheduling. Accesses to page $p_i$ introduce competing cache lines. More frequent accesses therefore produce larger eviction footprints on $s_i$. By accumulating these footprints over the prefill phase, \textsc{SparLeak} estimates $c_i$ for each monitored page and recovers the cumulative SIMA trace, without recovering page contents or exact runtime addresses. While this measurement does not reveal exact access counts, it reliably captures relative page-level sparsity induced by sparse attention, which is sufficient for downstream sparsity reconstruction and inference attacks. 


\subsubsection{Decoding-Phase SIMA Trace Extraction} \label{sec:sparsity_decoding}
During the decoding phase, the LLM generates the response autoregressively, producing one token per decoding step. At step $t$, LLM computes attention between the newly generated token and a subset of prompt tokens. We characterize the decoding-phase sparsity as a \emph{step-wise SIMA trace} $\{b_{1}^{t}, b_{2}^{t}, \ldots, b_{N}^{t}\}$, where $b_{i}^{t} \in \{0,1\}$ indicates whether the KV residing in the page $p_i$ are accessed by sparse attention at step $t$. Unlike the cumulative trace in the prefill phase, this binary trace captures per-step page participation and evolves across decoding steps.


\noindent\textbf{Attack Implementation.} \textsc{SparLeak} exploits the GPU Translation Lookaside Buffer (TLB), which caches virtual-to-physical page translations and exposes page-level access footprints under contention. Our design utilizes the GPU TLB \textsc{Evict} + \textsc{Reload} primitive demonstrated in prior work~\cite{zhang2023t}. The key idea is to infer whether a monitored prompt-side page $p_i$ is accessed during the decoding phase by measuring the translation latency of an attacker-owned probe page $q_i$ that conflicts with $p_i$ in the TLB, without recovering virtual-page contents or exact victim addresses. Under the fixed model and runtime configuration, the victim pages remain stable across invocations. The spy therefore calibrates two components offline: for each monitored page $p_i$, it allocates candidate pages in its own address space and selects a probe page $q_i$ whose translation conflicts with the same or a highly overlapping TLB set as $p_i$, following the TLB-set calibration methodology of~\cite{zhang2023t}. The spy also estimates the average decoding-step duration from the periodicity of offline probe traces and sets the probing-window length $\Delta$ to approximately cover one generation step. In our implementation, one probing round completes within $\Delta$, enabling step-granularity observation.

\begin{figure}[t]
    \centering
    \includegraphics[width=\linewidth]{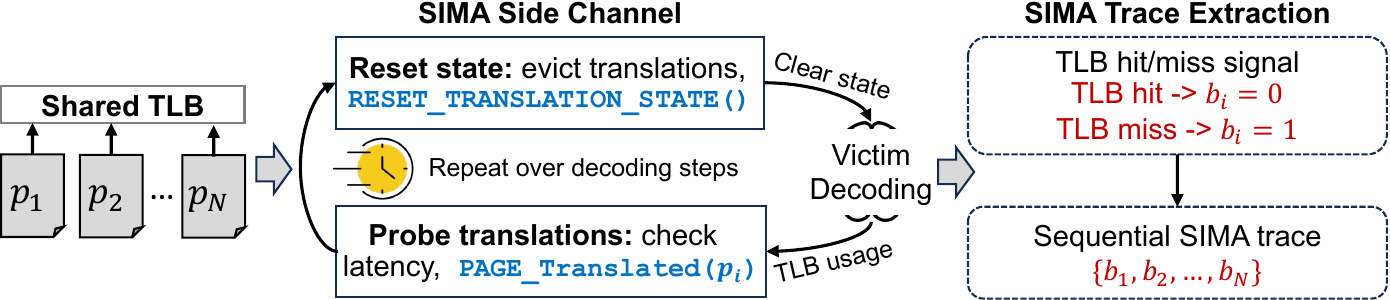}
    \caption{Spy process for recovering the sequential page-level sparsity pattern during the decoding phase. At each decoding step, the spy detects whether KV data from a prompt page are accessed by checking the persistence of its address translation state.}
    \label{fig:decoding_attack}
\end{figure}

At runtime, for each window, the spy prepares the calibrated probe-page translations, allows the victim to execute during the window, and then reloads each $q_i$ to measure its translation latency, as shown in \autoref{fig:decoding_attack}. If sparse attention accesses $p_i$, the resulting TLB contention may displace the translation entry of $q_i$, causing a high-latency reload. \textsc{SparLeak} therefore sets $b_i^t=1$ when the measured latency exceeds a calibrated threshold, and otherwise sets $b_i^t=0$. Repeating this procedure across monitored pages yields an approximate step-wise SIMA trace $\{b_{1}^{t}, \ldots, b_{N}^{t}\}$.

In practice, a window of length $\Delta$ may not perfectly align with a single decoding step due to scheduling jitter or other noise. To tolerate such misalignment, \textsc{SparLeak} aggregates measurements from adjacent probing windows and applies majority voting to suppress transient noise. 

\section{Sparsity Pattern Reconstruction}\label{sec:sparsity_reconstruction}
\begin{figure}[t]
    \centering
    \includegraphics[width=\linewidth]{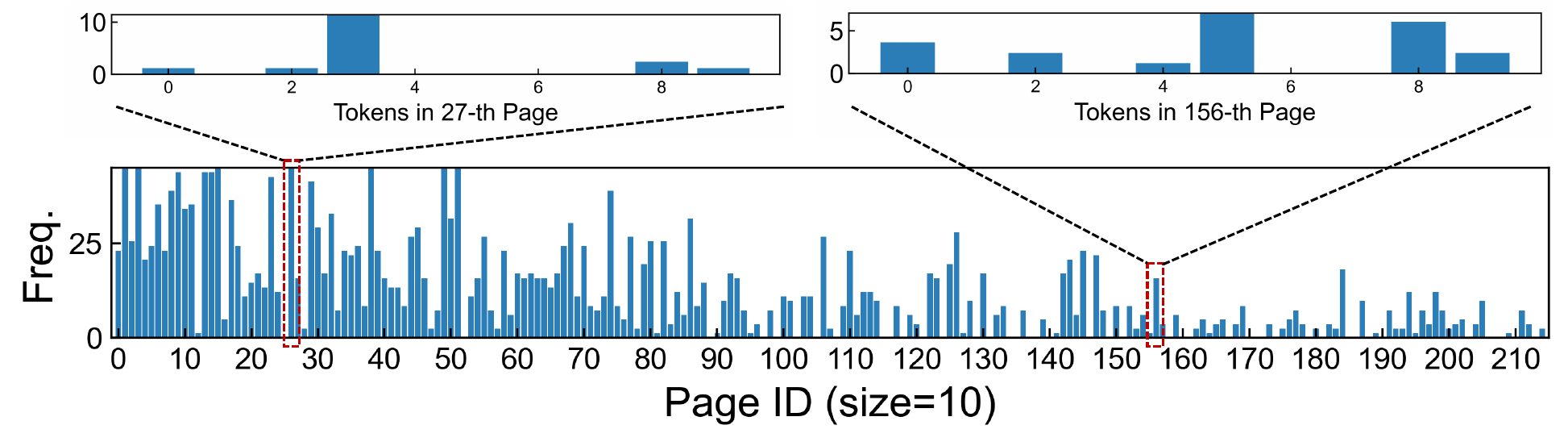}
    \caption{Illustration of page-level and token-level sparsity patterns in the prefill phase.}
    \label{fig:p2t_distribution}
\end{figure}


This section presents reconstruction methods for refining token-level sparsity from extracted SIMA traces, which provide more fine-grained information for subsequent attacks.

\subsection{Prefill-Phase Sparsity Reconstruction}
Reconstructing token-level sparsity patterns from page-level SIMA traces is challenging because page-level observations compress the activations of multiple tokens into a single value, thereby losing fine-grained intra-page information. A straightforward reconstruction that uniformly distributes the observed accesses across tokens within each page would fail to preserve the highly non-uniform structure of sparse attention. Fortunately, we find that the page-level cumulative access frequency is correlated with the latent token-level activation distribution within the page, providing a useful statistical cue for reconstruction.

To demonstrate this correlation, \autoref{fig:p2t_distribution} presents page-level aggregations of prefill-phase sparsity patterns with a page size of 10 tokens. The bottom panel shows the page-level cumulative SIMA trace, while the top panels show the token-level activation frequencies within two representative pages. A clear correlation emerges: pages with higher cumulative frequency tend to exhibit more skewed, long-tailed token-level activation patterns, whereas low-frequency pages show flatter and more uniform activation across tokens. For example, the 27-th page (top-left) lies in a high-frequency region and contains a small number of tokens that dominate activation, while 156-th page (top-right) lies in a low-frequency region and displays an even activation distribution.

This correlation is consistent with how sparse attention selects tokens during the prefill phase. For all query tokens, sparse attention repeatedly reuses only a small subset of highly relevant prompt tokens, while most prompt tokens are selected rarely~\cite{xiaoduoattention, zhu2025sampleattention}. Such repeated selection yields a long-tailed distribution of token-level activation frequencies. Since page-level frequency is computed by summing token-level activations within each page, pages that contain one or more frequently selected tokens naturally accumulate higher total frequency and appear more concentrated at the token level. Conversely, pages dominated by infrequently selected tokens accumulate lower frequency and consequently appear flatter.

Motivated by this observation, \textsc{SparLeak} performs distributional reconstruction to reconstruct an approximate token-level sparsity from page-level SIMA traces. Concretely, \textsc{SparLeak} treats each page’s observed frequency as a control signal that determines how concentrated the token-level sparsity distribution should be: higher-frequency pages are modeled with more skewed token participation, while lower-frequency pages are modeled with flatter participation. The reconstructed token-level sparsity is generated in a way that preserves consistency with observed page-level frequency. 

\subsection{Decoding-Phase Sparsity Reconstruction}
\begin{figure}[t]
    \centering
    \includegraphics[width=\linewidth]{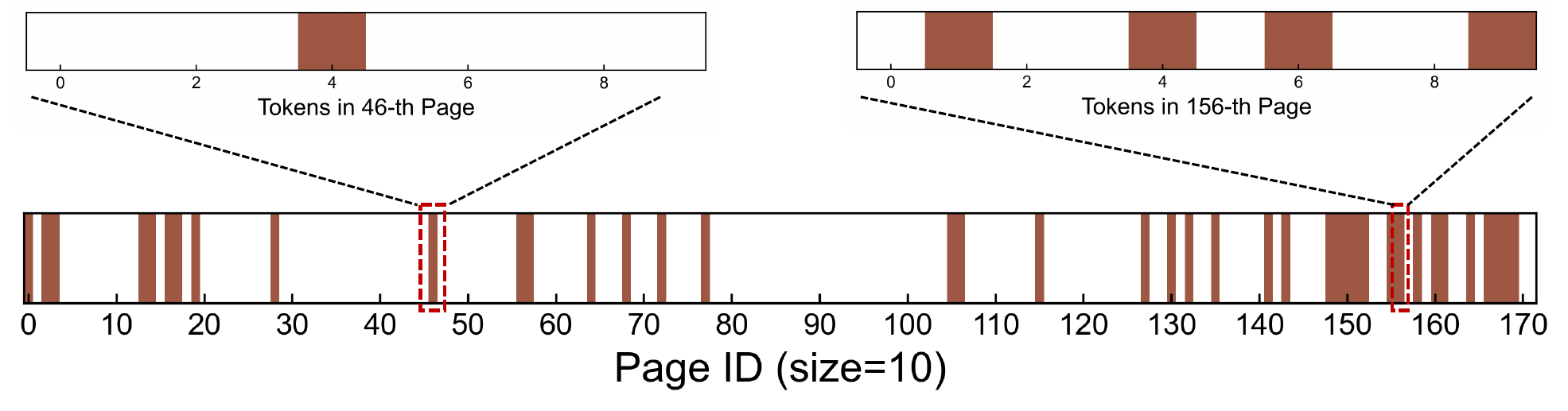}
    \caption{Illustration of page-level and token-level sparsity patterns in the decoding phase.}
    \label{fig:decoding_p2t_distribution}
\end{figure}
The decoding phase introduces a distinct reconstruction challenge. Unlike the cumulative prefill-phase trace, the step-wise SIMA trace provides only a binary observation for each page: an activated page indicates that some token within it is accessed, but does not reveal its intra-page activation pattern. We therefore examine whether the spatial distribution of activated pages at each decoding step provides additional information for reconstruction.

\autoref{fig:decoding_p2t_distribution} illustrates the relationship between page-level and token-level sparsity during the decoding phase for a representative decoding step (step~5 in this example). We partition the prompt into pages of 10 contiguous tokens and visualize two views: (i) the step-wise page-level SIMA trace observed at decoding time (bottom), and (ii) the corresponding ground-truth token-level activations within two selected pages (top), which are shown only for illustration. An empirical correlation can be observed: pages that are densely activated at the page level during the decoding phase tend to contain multiple activated tokens, whereas pages that are sparsely activated typically contain only a small number of active tokens. For example, the page region around the 156-th page exhibits frequent page-level activation and contains several activated tokens, while the region around the 46-th page is rarely activated and shows only limited token-level activity. Although the exact activation patterns vary across decoding steps and individual tokens, the number of active tokens within a page closely follows the local density of the page-level SIMA trace.

This behavior is consistent with the locality of sparse attention during the decoding phase. At each decoding step, sparse attention selects only a small set of prompt tokens, and these selections are not uniformly distributed across the prompt. Instead, selected tokens tend to cluster within a few contiguous regions, while most regions receive no activation. The page-level SIMA trace provides a coarse projection of this token-level selection onto page boundaries: clustered token selections lead to dense page-level activation, whereas isolated selections result in sparse page-level signals.

Motivated by this observation, \textsc{SparLeak} reconstructs an approximate token-level sparsity pattern by exploiting spatial locality in sparse attention. \textsc{SparLeak} uses the local density of page-level activations around a target page as a proxy for the extent of intra-page token participation. Based on this proxy, \textsc{SparLeak} instantiates a representative token-level sparsity pattern that is consistent with the observed page-level SIMA trace, while preserving the spatial structure and step-wise dynamics of sparse attention. 
\section{Sparsity-Guided Inference Attack}\label{sec:inference_attack}
In this section, we describe how \textsc{SparLeak} infers sensitive information from both user queries and LLM responses.

\subsection{Query Attribute Inference}
The adversary aims to infer sensitive attributes of a user query, such as health-related semantic categories, consistent with prior inference attacks~\cite{pan2020privacy, lyu2020differentially}. \textsc{SparLeak} formulates this as supervised classification using reconstructed token-level sparsity profiles from the prefill phase (\S\ref{sec:sparsity_reconstruction}), which capture attention allocation across prompt tokens. During offline profiling, the adversary collects sparsity profiles paired with known attribute labels from its own queries. The classifier learns associations between these structural patterns and semantic attributes, enabling online inference without access to the victim's query text.



\noindent\textbf{Normalization for scale invariance.}
Token-level sparsity profiles are affected by query length and the overall amount of attention activity. To prevent these factors from obscuring attribute-specific patterns, we normalize each sparsity profile to remove global scale differences. This emphasizes the relative distribution of sparse attention across prompt tokens, allowing the classifier to focus on structural variations that are more indicative of semantic attributes.

\noindent\textbf{Cross-layer and cross-head feature composition.}
Sparse attention decisions are made independently across layers and attention heads, and different components may capture complementary aspects of the input. Rather than collapsing these signals prematurely, we retain sparsity features from all layers and heads and concatenate them into a unified representation. This preserves cross-layer regularities and variability in sparsity behavior, improving the robustness and accuracy of attribute inference.

\subsection{Autoregressive Token Recovery}
\textsc{SparLeak} recovers generated responses token by token using step-wise token-level sparsity profiles reconstructed during decoding (\S\ref{sec:sparsity_decoding}). We formulate recovery as classification at each decoding step. During offline profiling, the adversary pairs step-wise sparsity profiles with known output tokens from its own queries. Features from the monitored layers and KV heads are concatenated to train a token classifier. Online recovery applies this classifier sequentially to reconstructed profiles without accessing victim outputs.


\noindent\textbf{Sequential feature augmentation.}
Although sparsity profiles at individual decoding steps already contain discriminative information, autoregressive generation exhibits strong temporal dependence. To capture such dependencies without increasing feature dimensionality, we augment each step’s feature representation with aggregated sparsity information from preceding decoding steps. This allows the classifier to exploit temporal regularities in how sparsity patterns evolve as decoding progresses.

\noindent\textbf{Restricted token prediction space.}
To reduce the complexity and class imbalance of full-vocabulary classification, \textsc{SparLeak} constructs a task-specific candidate set from the offline profiling corpus. The set is fixed before online execution and reused within the same task. Online recovery uses SIMA traces to predict tokens within this set, without accessing victim outputs. Supplementary experiments examine whether the gains come from SIMA or the restricted candidate set.
\section{Evaluation}

\begin{table*}[t]
\centering
\resizebox{\linewidth}{!}{
\begin{tabular}{c|c|ccc|ccc|ccc}
\toprule
\multirow{2}{*}{\textbf{Attribute}}
& \multirow{2}{*}{\textbf{Signal}}
& \multicolumn{3}{c|}{\textbf{LongChat-7B}~\cite{li2023long}}
& \multicolumn{3}{c|}{\textbf{LLaMA3-8B}~\cite{grattafiori2024llama}}
& \multicolumn{3}{c}{\textbf{Qwen3-8B}~\cite{yang2025qwen3}}\\
\cline{3-11}
& & \textbf{Act.} & \textbf{Block} & \textbf{Learn.} & \textbf{Act.} & \textbf{Block} & \textbf{Learn.} & \textbf{Act.} & \textbf{Block} & \textbf{Learn.} \\
\toprule

\multirow[c]{4}{*}{\shortstack[c]{Illness\\(60)}}
& O(T) & 89.4$\pm$1.1 & 83.3$\pm$0.7 & 99.8$\pm$0.1 & 86.9$\pm$0.8 & 92.1$\pm$1.1 & 99.5$\pm$0.4 & 86.3$\pm$0.8 & 86.8$\pm$0.5 & 98.4$\pm$0.3 \\
& O(P+R) & 87.3$\pm$0.7 & 81.7$\pm$1.1 & 99.4$\pm$0.3 & 84.5$\pm$0.5 & 90.9$\pm$0.6 & 99.2$\pm$0.1 & 84.4$\pm$1.0 & 85.2$\pm$0.2 & 98.1$\pm$0.3 \\
& R(P) & 70.4$\pm$2.6 & 66.4$\pm$1.2 & 78.7$\pm$0.7 & 67.1$\pm$1.9 & 66.51$\pm$1.4 & 74.4$\pm$1.2 & 60.3$\pm$3.6 & 62.3$\pm$1.8 & 75.3$\pm$1.0 \\
& \textbf{R(P+R)}
& \textbf{85.3$\pm$0.4} & \textbf{80.5$\pm$0.3} & \textbf{98.8$\pm$0.6} & \textbf{84.1$\pm$0.6} & \textbf{88.4$\pm$0.7} & \textbf{98.7$\pm$0.5} & \textbf{81.8$\pm$0.4} & \textbf{83.0$\pm$0.5} & \textbf{94.7$\pm$0.2} \\
\midrule

\multirow[c]{4}{*}{\shortstack[c]{Age\\(8)}}
& O(T) & 86.2$\pm$0.5 & 79.7$\pm$0.3 & 99.9$\pm$0.1 & 68.4$\pm$1.0 & 81.3$\pm$0.8 & 99.4$\pm$0.3 & 99.5$\pm$0.4 & 99.1$\pm$0.2 & 99.8$\pm$0.1 \\
& O(P+R) & 85.3$\pm$0.2 & 77.3$\pm$0.4 & 99.8$\pm$0.1 & 67.3$\pm$0.5 & 79.2$\pm$0.4 & 98.2$\pm$0.2 & 99.4$\pm$0.1 & 98.1$\pm$0.2 & 99.6$\pm$0.1 \\
& R(P) & 68.3$\pm$1.2 & 58.0$\pm$2.0 & 82.3$\pm$1.8 & 45.2$\pm$3.3 & 59.7$\pm$1.4 & 81.3$\pm$0.8 & 79.8$\pm$1.4 & 81.5$\pm$2.2 & 85.1$\pm$1.4 \\
& \textbf{R(P+R)}
& \textbf{82.3$\pm$0.4} & \textbf{74.5$\pm$0.4} & \textbf{99.8$\pm$0.1} & \textbf{62.0$\pm$0.9} & \textbf{77.9$\pm$0.5} & \textbf{97.1$\pm$0.3} & \textbf{97.4$\pm$0.1} & \textbf{97.1$\pm$0.3} & \textbf{99.4$\pm$0.1} \\
\midrule

\multirow[c]{4}{*}{\shortstack[c]{Gender\\(2)}}
& O(T) & 98.2$\pm$0.2 & 99.1$\pm$0.3 & 99.8$\pm$0.1 & 94.1$\pm$0.3 & 99.9$\pm$0.0 & 99.7$\pm$0.2 & 99.5$\pm$0.4 & 99.9$\pm$0.1 & 100.0$\pm$0.0 \\
& O(P+R) & 97.5$\pm$0.3 & 98.7$\pm$0.3 & 99.4$\pm$0.3 & 91.8$\pm$0.4 & 99.8$\pm$0.2 & 99.6$\pm$0.1 & 99.5$\pm$0.2 & 99.6$\pm$0.3 & 100.0$\pm$0.0 \\
& R(P) & 81.5$\pm$0.6 & 82.3$\pm$0.4 & 83.5$\pm$1.3 & 77.3$\pm$1.3& 80.6$\pm$1.5 & 83.6$\pm$1.6 & 85.2$\pm$1.6 & 84.7$\pm$2.1 & 85.4$\pm$2.8 \\
& \textbf{R(P+R)}
& \textbf{96.4$\pm$0.2} & \textbf{96.8$\pm$0.4} & \textbf{99.3$\pm$0.6} & \textbf{90.6$\pm$0.4} & \textbf{99.7$\pm$0.2} & \textbf{99.4$\pm$0.5} & \textbf{99.3$\pm$0.3} & \textbf{99.7$\pm$0.1} & \textbf{100.0$\pm$0.0} \\
\midrule

\multirow[c]{4}{*}{\shortstack[c]{Blood\\(8)}}
& O(T) & 72.3$\pm$0.6 & 93.5$\pm$0.3 & 99.8$\pm$0.2 & 60.4$\pm$0.6 & 95.3$\pm$0.5 & 99.9$\pm$0.0 & 81.4$\pm$0.6 & 99.5$\pm$0.4 & 97.4$\pm$0.3 \\
& O(P+R) & 71.1$\pm$0.4 & 92.6$\pm$0.3 & 99.8$\pm$0.1 & 59.2$\pm$0.4 & 93.3$\pm$0.4 & 99.9$\pm$0.1 & 80.3$\pm$0.4 & 99.5$\pm$0.3 & 97.2$\pm$0.2 \\
& R(P) & 49.3$\pm$1.1 & 77.5$\pm$0.9 & 82.2$\pm$1.6 & 37.6$\pm$5.5 & 78.7$\pm$0.6 & 85.6$\pm$2.5 & 60.5$\pm$1.5 & 81.6$\pm$1.6 & 80.3$\pm$1.7 \\
& \textbf{R(P+R)}
& \textbf{67.9$\pm$0.6} & \textbf{91.5$\pm$0.3} & \textbf{99.7$\pm$0.1} & \textbf{55.2$\pm$0.3} & \textbf{92.0$\pm$0.3} & \textbf{99.8$\pm$0.1} & \textbf{77.4$\pm$0.5} & \textbf{99.1$\pm$0.4} & \textbf{95.5$\pm$0.2} \\
\toprule

\multirow[c]{4}{*}{\shortstack[c]{Entity\\(70)}}
& O(T) & 86.4$\pm$0.3 & 91.3$\pm$0.2 & 93.4$\pm$0.4 & 84.3$\pm$0.2 & 86.1$\pm$0.5 & 89.3$\pm$0.2 & 91.9$\pm$0.3 & 95.4$\pm$0.3 & 97.2$\pm$0.4 \\
& O(P+R) & 85.0$\pm$0.3 & 90.3$\pm$0.4 & 92.3$\pm$0.3 & 84.0$\pm$0.2 & 84.3$\pm$0.2 & 88.4$\pm$0.2 & 90.7$\pm$0.2 & 93.3$\pm$0.3 & 96.4$\pm$0.3 \\
& R(P) & 68.1$\pm$1.3 & 71.4$\pm$1.1 & 79.2$\pm$0.9 & 66.6$\pm$1.1 & 69.4$\pm$1.0 & 68.5$\pm$1.2 & 70.4$\pm$1.8 & 80.4$\pm$0.5 & 82.1$\pm$2.3 \\
& \textbf{R(P+R)}
& \textbf{84.0$\pm$0.3} & \textbf{88.3$\pm$0.2} & \textbf{90.3$\pm$0.4} & \textbf{80.3$\pm$0.5} & \textbf{82.0$\pm$0.3} & \textbf{84.9$\pm$0.3} & \textbf{88.6$\pm$0.3} & \textbf{92.2$\pm$0.3} & \textbf{94.8$\pm$0.4} \\
\midrule

\multirow[c]{4}{*}{\shortstack[c]{Period\\(4)}}
& O(T) & 94.0$\pm$0.4 & 94.3$\pm$0.2 & 97.4$\pm$0.3 & 95.9$\pm$0.4 & 98.3$\pm$0.2 & 99.1$\pm$0.2 & 97.7$\pm$0.3 & 98.5$\pm$0.1 & 99.8$\pm$0.1 \\
& O(P+R) & 93.2$\pm$0.3 & 92.2$\pm$0.2 & 97.0$\pm$0.3 & 95.1$\pm$0.2 & 97.4$\pm$0.2 & 98.3$\pm$0.3 & 96.1$\pm$0.3 & 97.7$\pm$0.2 & 99.7$\pm$0.1 \\
& R(P) & 79.6$\pm$1.1 & 77.9$\pm$1.0 & 81.2$\pm$1.3 & 80.5$\pm$0.8 & 82.4$\pm$1.7 & 83.2$\pm$1.6 & 81.4$\pm$1.5 & 79.9$\pm$2.5 & 84.3$\pm$1.4 \\
& \textbf{R(P+R)}
& \textbf{92.7$\pm$0.2} & \textbf{91.5$\pm$0.4} & \textbf{95.8$\pm$0.3} & \textbf{93.6$\pm$0.2} & \textbf{96.6$\pm$0.2} & \textbf{97.9$\pm$0.2} & \textbf{95.8$\pm$0.2} & \textbf{96.5$\pm$0.3} & \textbf{99.4$\pm$0.4} \\
\midrule

\multirow[c]{4}{*}{\shortstack[c]{Topic\\(5)}}
& O(T) & 93.4$\pm$0.3 & 97.3$\pm$0.2 & 98.6$\pm$0.2 & 91.1$\pm$0.3 & 95.3$\pm$0.4 & 95.3$\pm$0.4 & 89.0$\pm$0.2 & 91.8$\pm$0.3 & 92.5$\pm$0.5 \\
& O(P+R) & 92.6$\pm$0.3 & 96.9$\pm$0.1 & 96.2$\pm$0.3 & 90.1$\pm$0.3 & 95.0$\pm$0.4 & 94.3$\pm$0.3 & 87.3$\pm$0.2 & 90.2$\pm$0.4 & 91.3$\pm$0.2 \\
& R(P) & 76.3$\pm$1.2 & 79.5$\pm$1.3 & 81.2$\pm$2.2 & 72.3$\pm$1.3 & 78.2$\pm$2.1 & 79.4$\pm$2.3 & 75.5$\pm$2.1 & 76.3$\pm$1.6 & 77.3$\pm$1.7 \\
& \textbf{R(P+R)}
& \textbf{90.3$\pm$0.4} & \textbf{94.0$\pm$0.3} & \textbf{94.8$\pm$0.4} & \textbf{87.8$\pm$0.4} & \textbf{92.2$\pm$0.4} & \textbf{92.4$\pm$0.3} & \textbf{85.8$\pm$0.2} & \textbf{88.2$\pm$0.3} & \textbf{89.8$\pm$0.2} \\
\toprule

\multirow[c]{4}{*}{\shortstack[c]{Domain\\(15)}}
& O(T) & 94.1$\pm$0.2 & 94.6$\pm$0.4 & 98.5$\pm$0.2 & 95.8$\pm$0.3 & 97.3$\pm$0.4 & 98.8$\pm$0.2 & 91.2$\pm$0.3 & 91.3$\pm$0.3 & 98.4$\pm$0.2 \\
& O(P+R) & 93.1$\pm$0.3 & 93.3$\pm$0.3 & 97.1$\pm$0.3 & 94.3$\pm$0.3 & 96.0$\pm$0.3 & 97.2$\pm$0.2 & 90.1$\pm$0.3 & 91.1$\pm$0.3 & 97.2$\pm$0.3 \\
& R(P) & 79.3$\pm$1.0 & 80.1$\pm$1.6 & 81.3$\pm$1.5 & 79.4$\pm$1.8 & 80.2$\pm$1.2 & 82.1$\pm$2.2 & 72.5$\pm$1.5 & 71.5$\pm$1.1 & 80.2$\pm$1.9 \\
& \textbf{R(P+R)}
& \textbf{90.2$\pm$0.6} & \textbf{90.8$\pm$0.4} & \textbf{95.3$\pm$0.4} & \textbf{92.3$\pm$0.4} & \textbf{94.9$\pm$0.3} & \textbf{95.6$\pm$0.3} & \textbf{87.6$\pm$0.4} & \textbf{87.8$\pm$0.4} & \textbf{95.2$\pm$0.2} \\
\midrule

\multirow[c]{4}{*}{\shortstack[c]{Role\\(8)}}
& O(T) & 97.9$\pm$0.2 & 97.4$\pm$0.4 & 99.2$\pm$0.2 & 99.8$\pm$0.1 & 99.3$\pm$0.2 & 99.8$\pm$0.1 & 95.9$\pm$0.3 & 98.3$\pm$0.3 & 98.3$\pm$0.2 \\
& O(P+R) & 96.3$\pm$0.2 & 95.8$\pm$0.3 & 98.4$\pm$0.2 & 99.3$\pm$0.0 & 98.3$\pm$0.2 & 99.1$\pm$0.2 & 94.3$\pm$0.2 & 96.8$\pm$0.3 & 97.2$\pm$0.1 \\
& R(P) & 79.5$\pm$1.9 & 78.4$\pm$1.7 & 79.3$\pm$1.6 & 79.9$\pm$1.0 & 79.1$\pm$1.4 & 81.1$\pm$2.4 & 75.6$\pm$2.0 & 79.9$\pm$0.7 & 81.3$\pm$3.1 \\
& \textbf{R(P+R)}
& \textbf{94.9$\pm$0.3} & \textbf{94.8$\pm$0.2} & \textbf{96.8$\pm$0.3} & \textbf{97.6$\pm$0.3} & \textbf{96.9$\pm$0.3} & \textbf{97.8$\pm$0.4} & \textbf{92.6$\pm$0.3} & \textbf{95.4$\pm$0.2} & \textbf{95.1$\pm$0.3} \\
\midrule

\multirow[c]{4}{*}{\shortstack[c]{Location\\(20)}}
& O(T) & 88.8$\pm$0.1 & 89.6$\pm$0.3 & 92.2$\pm$0.3 & 87.7$\pm$0.2 & 85.4$\pm$0.2 & 89.6$\pm$0.3 & 91.8$\pm$0.2 & 93.3$\pm$0.3 & 95.2$\pm$0.3 \\
& O(P+R) & 87.3$\pm$0.2 & 87.3$\pm$0.4 & 91.4$\pm$0.2 & 86.6$\pm$0.3 & 84.3$\pm$0.3 & 88.0$\pm$0.2 & 90.3$\pm$0.3 & 92.1$\pm$0.3 & 93.9$\pm$0.3 \\
& R(P) & 69.3$\pm$1.6 & 70.3$\pm$1.3 & 71.5$\pm$2.0 & 69.8$\pm$1.2 & 68.5$\pm$1.6 & 71.4$\pm$1.7 & 72.5$\pm$1.8 & 73.1$\pm$1.3 & 75.2$\pm$1.4 \\
& \textbf{R(P+R)}
& \textbf{85.5$\pm$0.3} & \textbf{86.9$\pm$0.4} & \textbf{88.1$\pm$0.2} & \textbf{84.9$\pm$0.2} & \textbf{82.3$\pm$0.4} & \textbf{86.1$\pm$0.4} & \textbf{88.3$\pm$0.4} & \textbf{89.7$\pm$0.3} & \textbf{91.9$\pm$0.3} \\

\bottomrule
\end{tabular}
}
\caption{Query attribute inference PASR (\%) on Healthcare~\cite{health} (Illness, Age, Gender, Blood), Financial QA~\cite{financialqa} (Entity, Period, Topic), and Legal QA~\cite{legalqa} (Domain, Role, Location). Each entry report the mean $\pm$ 95\% confidence interval over 20 runs.
}
\label{tab:prefill_attack_results}
\end{table*}

\begin{figure*}
    \centering
    \includegraphics[width=\linewidth]{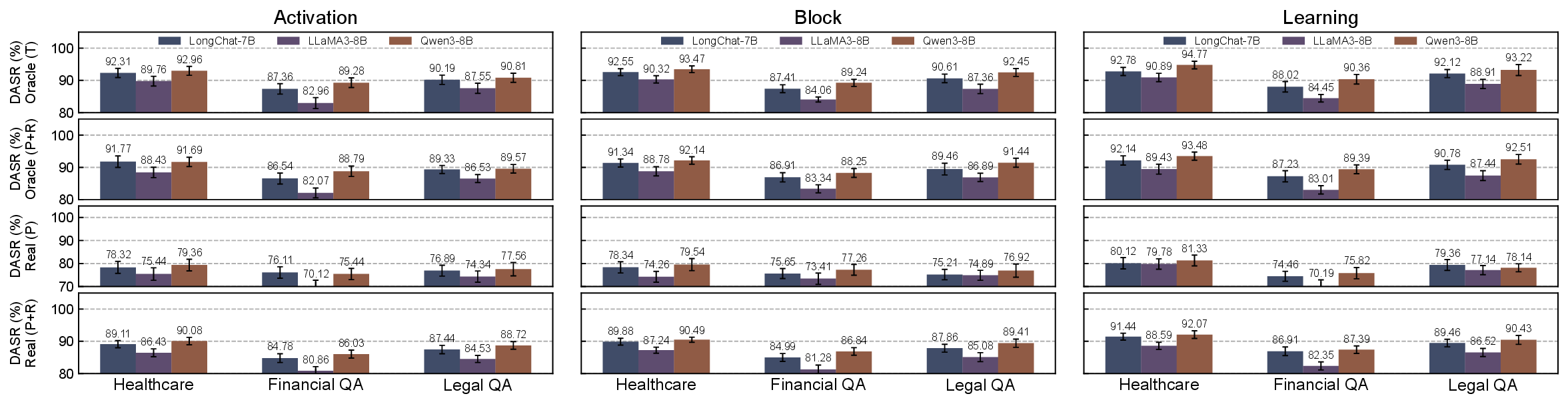} 
    \caption{Performance of autoregressive token recovery attacks, measured by DASR.}
    \label{fig:overall_decoding}
\end{figure*}

\subsection{Evaluation Setup}
\noindent\textbf{Models.} We evaluate three representative LLMs from different model families: LongChat-7B~\cite{li2023long}, LLaMA3-8B~\cite{grattafiori2024llama}, and Qwen3-8B~\cite{yang2025qwen3}. Each model is configured with three classes of sparse attention mechanisms: (1) \emph{activation-based saliency}~\cite{zhang2023h2o, xu2025xattention, cho2024sparc}, which identifies important tokens by comparing intermediate activations; (2) \emph{block-level scoring}~\cite{desaihashattention, tang2024quest}, which groups tokens into blocks or pages and computes approximate attention scores at the block level; and (3) \emph{learned predictors}~\cite{treviso2022predicting, gao2024seerattention, xu2025specontext}, which employ lightweight predictors to directly select salient tokens.

\noindent\textbf{Datasets.} We use three widely adopted datasets for privacy-sensitive evaluation: Healthcare~\cite{health}, Financial~QA~\cite{financialqa}, and Legal~QA~\cite{legalqa}. In the Healthcare dataset, each sample contains only attribute values. We construct LLM inference requests by transforming these attributes into unstructured queries using templates generated by a public model. For the Financial~QA and Legal~QA datasets, we summarize multiple attributes within each query to enable attribute inference attacks. All constructed queries are fed into the deployed LLMs to generate responses, which are subsequently used to perform autoregressive token recovery attacks.

\noindent\textbf{Metrics.} We evaluate attack effectiveness separately for the prefill and decoding phases using phase-specific attack success rates. The prefill-phase attack success rate (PASR) is the fraction of correct attribute predictions among queries containing the corresponding attribute. Its denominator may therefore vary across attributes. The decoding-phase attack success rate (DASR) is the fraction of correctly reconstructed tokens per response, averaged across queries. Ablation studies additionally report the similarity between extracted SIMA traces and ground-truth sparsity patterns.

\noindent\textbf{Environment.} We evaluate the performance of \textsc{SparLeak} on an NVIDIA L20 GPU (48GB) under Ubuntu 20.04 (Linux 5.15.0), using NVIDIA driver 550.67, CUDA 11.8, PyTorch 2.6.0.

\subsection{Attack Performance}\label{sec:overall_performance}

\autoref{tab:prefill_attack_results} and \autoref{fig:overall_decoding} report PASR and DASR, respectively. The results distinguish token-level oracle patterns (O(T)), reconstructed oracle page patterns (O(P+R)), raw page-level observations (R(P)), and reconstructed real observations (R(P+R)), where R(P+R) represents the complete end-to-end attack.

\noindent\textbf{Overall effectiveness.} In the end-to-end attack, average PASR exceeds 91\% across all datasets, while DASR ranges from 84.6\% to 89.4\%, with every setting exceeding 80\%. The high O(T) performance confirms that token-level sparsity patterns inherently encode substantial information about private queries and generated responses. O(P+R) remains within approximately one percentage point of O(T), showing limited information loss from page aggregation after reconstruction. End-to-end performance is only about 2.5 points below the token-level oracle, showing that reconstruction recovers most information from noisy page-level traces.



\noindent\textbf{Impact of attribute complexity.} PASR varies across attribute types. Attributes with fewer classes, such as \textit{Gender} in Healthcare, achieve near-perfect accuracy (around 97.9\%), whereas those with larger label spaces, such as \textit{Entity} in Financial QA, yield lower PASR (approximately 87.3\%), reflecting the difficulty of distinguishing more semantic classes. Decoding DASR is less sensitive to attribute semantics because token recovery reconstructs output sequences. However, Financial QA consistently yields lower DASR than Healthcare and Legal QA, reflecting the greater lexical diversity and structural complexity of financial responses.

\noindent\textbf{Impact of sparse attention mechanisms.} Sparse attention mechanisms affect attack performance in both phases. Activation-based methods yield lower prefill PASR (approximately 86.4\%) than block-level scoring and learned predictors, with the latter achieving the highest PASR (around 95.5\%). Decoding follows a similar trend: activation-based methods yield lower DASR, block-level scoring provides moderate improvements, and learned predictors perform best across datasets and model backbones. These results suggest that more accurate sparsity estimation reduces information loss in SIMA traces, benefiting both attribute inference and token recovery.

\noindent\textbf{Impact of LLM architectures.} \textsc{SparLeak} achieves average prefill PASRs of approximately 90.6\%, 89.4\%, and 92.7\% on LongChat-7B, LLaMA3-8B, and Qwen3-8B, respectively. In addition, Qwen3-8B consistently achieves the highest decoding DASR across datasets (\autoref{fig:overall_decoding}). This consistency suggests that its sparsity patterns across layers and heads are more stable and distinctive, yielding cleaner SIMA signals for attribute inference and token recovery.

\subsection{Ablation Study}\label{sec:ablation_study}
\begin{figure}[t]
    \centering
    \includegraphics[width=\linewidth]{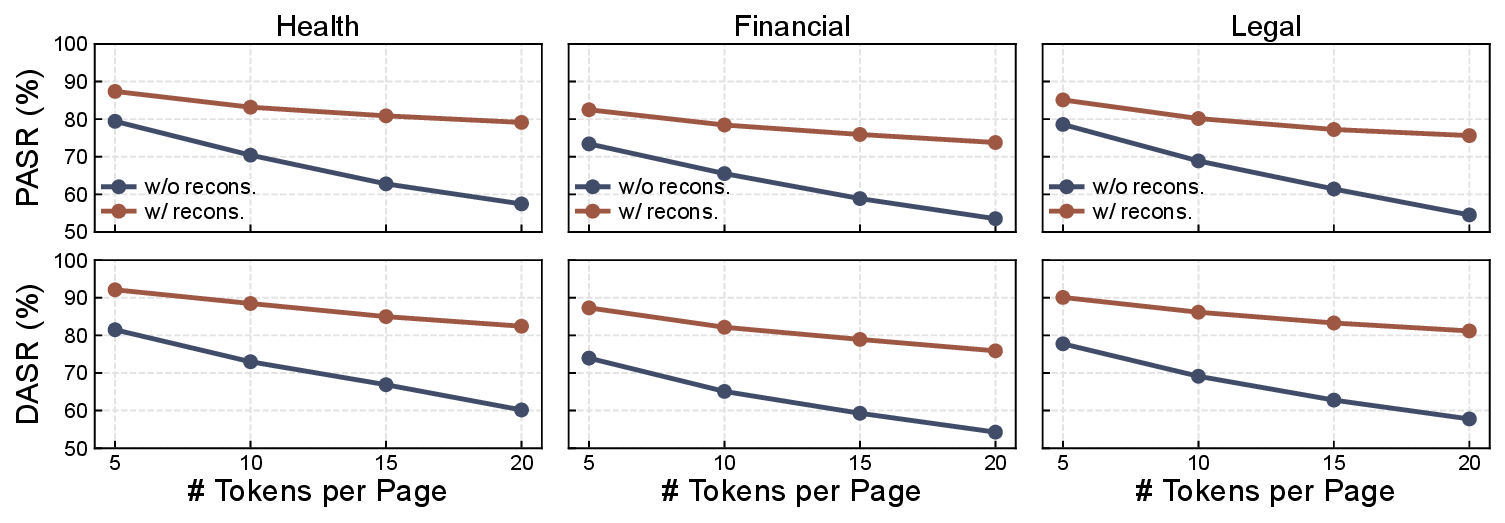} 
    \caption{The performance of sparsity reconstruction.}
    \label{fig:abl_reconstruction_all}
\end{figure}

We evaluate the proposed reconstruction methods by aggregating ground-truth token-level sparsity patterns into pages of 5 to 20 tokens, reflecting practical KV page granularities, and measuring PASR and DASR before and after reconstruction. As shown in \autoref{fig:abl_reconstruction_all}, coarser page aggregation increasingly obscures token-level information and degrades both attacks without reconstruction. At a page size of 20, the Healthcare PASR drops to 57.45\% without reconstruction but remains substantially higher when reconstruction is enabled. More ablation studies can be found at the submitted supplementary material.

\subsection{Practicality Analysis}
We further study the feasibility of \textsc{SparLeak}, which depends on (i) whether the offline preparation incurs prohibitive cost, and (ii) whether the online side-channel observations can be conducted unobtrusively in realistic deployment settings with strong stealthiness.

\noindent\textbf{Profiling settings and overhead.} 
Our primary evaluation assumes that the model checkpoint, tokenizer, sparse-attention implementation, and configuration are public. To isolate SIMA leakage from configuration mismatch, attacker runs an instrumented replica on controlled inputs to collect ground-truth sparsity patterns and train predictors. This profiling may run on a separate GPU, while page-probe calibration remains target-specific. We use 5,000 QAI queries per model and 3,000/3,000/2,000 ATR responses for LongChat-7B, LLaMA3-8B, and Qwen3-8B, respectively, with an 8:1:1 sample-disjoint split and task-specific 18-/34-layer residual predictors. As summarized in \autoref{tab:profiling_overhead}, LLM inference dominates the profiling cost: QAI requires one prefill pass per query, whereas ATR records step-wise patterns throughout decoding and therefore scales with response length. Predictor training adds negligible overhead, and this one-time cost is amortized across attacks using same configuration.


When model weights or sparse-attention configurations are unavailable, \textsc{SparLeak} instead performs target-interactive profiling. The attacker submits chosen queries while a co-resident spy records the corresponding real page-level SIMA traces; attacker-known attributes and returned tokens provide the QAI and ATR labels. Training directly on these trace--label pairs achieves average PASR and DASR of 90.97\% and 86.65\%, with maximum reductions of only 1.28 and 1.54 percentage points relative to replica-based profiling. Thus, \textsc{SparLeak} remains effective without model or sparsity access, at the cost of one-time same-GPU chosen-query collection.

\begin{table}[t]
\centering
\small
\resizebox{\linewidth}{!}{
\begin{tabular}{c|ccc|cccc}
\hline
\multirow{2}{*}{\makecell[c]{\textbf{Model}}}
& \multicolumn{3}{c|}{\textbf{Attribute Inference Attack}}
& \multicolumn{4}{c}{\textbf{Token Recovery Attack}} \\
\cline{2-8}
& \makecell[c]{Per-query\\prefill time}
& \makecell[c]{\#Samples}
& \makecell[c]{Cost}
& \makecell[c]{Per-token\\decode time}
& \makecell[c]{Avg.\\length}
& \makecell[c]{\#Samples}
& \makecell[c]{Cost} \\
\hline
LongChat-7B~\cite{li2023long} & 0.62s & 5,000 & \cellcolor{gray!15} 0.86h & 0.05s & 300 & 3,000 & 12.7h\cellcolor{gray!15} \\
LLaMA3-8B~\cite{grattafiori2024llama} & 0.58s & 5,000 & \cellcolor{gray!15} 0.82h & 0.06s & 300 & 3,000 & 15.2h\cellcolor{gray!15} \\
Qwen3-8B~\cite{yang2025qwen3} & 0.79s & 5,000 & \cellcolor{gray!15} 1.10h & 0.07s & 500 & 2,000 & \cellcolor{gray!15} 19.5h\\
\hline
\end{tabular}
}
\caption{Offline profiling overhead for query attribute inference and autoregressive token recovery attacks.
}
\label{tab:profiling_overhead}
\end{table}



\noindent\textbf{Stealthiness.} We assume an adversary co-located with the victim on the same physical GPU that passively collects side-channel signals during LLM inference. For \textsc{SparLeak} to be practical, such measurements must not noticeably perturb the victim’s execution. Our evaluation shows that this requirement is satisfied: enabling the spy process introduces only limited overhead, increasing inference latency by 1–3\% during the prefill phase (via L2-cache-based \textsc{Invalidate+Compare}) and 4–6\% during the decoding phase (via TLB-based \textsc{Evict} + \textsc{Reload}), with the worst-case slowdown remaining below 8\%. These overheads fall well within the normal latency variation of LLM serving environments, where scheduling and resource contention already induce comparable fluctuations. Consequently, the additional delay introduced by \textsc{SparLeak} is indistinguishable from benign system noise, rendering the attack stealthy in realistic deployment scenarios.

\subsection{Possible Mitigation}
In this section, we discuss potential defenses against the sparsity-induced side channels exploited by \textsc{SparLeak}, including model-level mitigations and general side-channel hardening. Their effectiveness must be considered alongside their impact on generation quality and serving efficiency. A direct mitigation reduces the information exposed by sparsity patterns through controlled, input-independent randomness. During inference, an $A\%$ fraction of tokens selected by importance scores is randomly replaced with other tokens in an input-independent manner. This perturbation disrupts the stability of input-dependent sparsity patterns, making SIMA traces less discriminative to an adversary. However, the replacement tokens may be less semantically relevant, weakening the selected context and degrading generation quality, particularly at high perturbation intensities.

We evaluate this mitigation on Financial~QA with perturbation intensities ranging from 10\% to 30\%. We report overall ASR, defined as the average of PASR and DASR, and generation quality measured by ROUGE. Increasing perturbation consistently reduces ASR but also degrades generation quality. For example, at 30\% perturbation, LLaMA3-8B's ROUGE score drops from 0.48 to 0.16. These results show that randomizing sparse selection mitigates leakage, but aggressive perturbation compromises the generation quality.


\begin{table}[t]
\centering
\small
\resizebox{\linewidth}{!}{
\begin{tabular}{c|cc|cc|cc|cc}
\hline
\multirow{2}{*}{\makecell[c]{\textbf{Intensity}}}
& \multicolumn{2}{c|}{\textbf{Baseline}}
& \multicolumn{2}{c|}{\textbf{10\%}} 
& \multicolumn{2}{c|}{\textbf{20\%}}
& \multicolumn{2}{c}{\textbf{30\%}}\\
\cline{2-9}
& \makecell[c]{ASR (\%)}
& \makecell[c]{Rouge}
& \makecell[c]{ASR (\%)}
& \makecell[c]{Rouge}
& \makecell[c]{ASR (\%)}
& \makecell[c]{Rouge}
& \makecell[c]{ASR (\%)}
& \makecell[c]{Rouge}\\
\hline
LongChat-7B~\cite{li2023long}  & 89.45 & 0.43 & 85.41 & 0.38 & 82.09 & 0.32 & 76.17 & 0.21 \\
LLaMA3-8B~\cite{grattafiori2024llama}  & 90.12 & 0.48 & 84.92 & 0.35 & 79.85 & 0.24 & 72.68 & 0.16 \\
Qwen3-8B~\cite{yang2025qwen3}  & 91.23 & 0.42 & 87.28 & 0.37 & 81.13 & 0.35 & 75.98 & 0.26 \\
\hline
\end{tabular}
}
\caption{Effectiveness of fixed sparsity pattern injection as a mitigation strategy. Mitigation intensity denotes the fraction of sparse tokens that are replaced by a random token set.}
\label{tab:mitgation}
\end{table}

Restricting fine-grained timers, monitoring interfaces, and memory-management primitives can hinder \textsc{SparLeak} by reducing trace fidelity without eliminating secret-dependent accesses. Fixed or page-padded schedules and dummy KV accesses reduce input dependence at the cost of sparse-attention efficiency. Tenant isolation and cache/TLB partitioning can prevent cross-tenant observation but require runtime or hardware support. Detecting repeated probing or anomalous cache/TLB activity can complement these defenses. Developing low-cost defenses that preserve both model quality and serving efficiency remains future work.


\section{Related Work}
\noindent\textbf{Privacy leakage in LLMs.}
Prior work has shown that privacy risks in LLMs arise from algorithm- or API-level exposure~\cite{das2025security, yan2024protecting}, including training-data memorization~\cite{liu2024precurious, wang2025leaner, akkus2025generated}, membership inference~\cite{wen2024membership, meeus2024did, he2025towards, das2025blind}, and prompt attribute inference~\cite{cao2025you, luo2025prompt, tan2025effectiveness, hui2024pleak,yang2025prsa}. These attacks typically rely on observing model outputs or interacting with the LLM through crafted prompts. In contrast, \textsc{SparLeak} targets inference-time execution behavior and demonstrates that sparse attention can externalize sensitive information through SIMA side channels in shared GPU components, without any interaction at the API level.

\noindent\textbf{Side-channel attacks on LLMs.}
Recent studies have revealed that LLM serving systems are vulnerable to various side channels, including timing-based attacks~\cite{zhang2024time, song2025early, wu2025know, zheng2024inputsnatch}, network traffic analysis~\cite{weiss2024your, mcdonald2025whisper}, and CPU cache monitoring~\cite{adiletta2025spill, gao2025know}. These approaches infer information through externally observable signals such as latency, encrypted traffic patterns, or embedding-table accesses. \textsc{SparLeak} differs fundamentally by exploiting memory access behavior intrinsic to sparse attention. This leakage is amplified by sparsity and manifests consistently across both prefill and decoding phases, enabling a unified framework for query attribute inference and autoregressive token recovery.

\section{Conclusion}
We present \textsc{SparLeak}, the first study of micro-architectural privacy leakage in sparse-attention-based LLM serving. Sparse attention produces distinctive sparsity-induced memory access (SIMA) patterns that expose sensitive information through shared GPU resources. \textsc{SparLeak} exploits phase-specific SIMA traces for query attribute inference during prefill and autoregressive token recovery during decoding. Evaluations across multiple LLMs, sparse attention mechanisms, and datasets demonstrate its effectiveness and robustness.







\bibliographystyle{ACM-Reference-Format}
\bibliography{ref}

\end{document}